\documentclass[conference]{IEEEtran}
\usepackage{geometry}
\usepackage{amsmath,amssymb,amsfonts}
\usepackage{algorithmic}
\usepackage{graphicx, subfigure}
\usepackage{multirow, booktabs, makecell, threeparttable, arydshln} 
\usepackage{textcomp}
\usepackage{xcolor}
\usepackage[hyphens]{url}
\def\BibTeX{{\rm B\kern-.05em{\sc i\kern-.025em b}\kern-.08em
    T\kern-.1667em\lower.7ex\hbox{E}\kern-.125emX}}
\usepackage{hyperref}
\usepackage[capitalise]{cleveref}
\usepackage[style=ieee,backend=biber,maxnames=6]{biblatex}
\DeclareMathOperator{\RMSNorm}{RMSNorm}

\usepackage{pgfplots}  
\usepgfplotslibrary{groupplots}  
\usepackage{xcolor}
\usepackage{tikz}  
\usetikzlibrary{
  shapes.geometric,
  arrows.meta,
  positioning,
  calc,
  fit,
  shadows,
  matrix
}
\pgfplotsset{compat=newest}

\colorlet{colRot}       {violet!80!black}
\colorlet{colSmooth}    {orange!80!black}
\colorlet{colSmoothRot}{colRot!50!colSmooth}

\tikzset{
  block/.style = {
    draw=black, thick,
    rectangle, rounded corners,
    minimum height=5em, minimum width=5em,
    align=center
  },
  nonlinearity/.style = {
     draw=black, thick,
    rectangle, rounded corners,
    minimum height=2.5em, minimum width=2.5em,
    align=center
  },
  smallblock/.style = {
    draw=black, thick,
    rectangle, rounded corners,
    minimum height=5em, minimum width=1em,
    align=center
  },
  operator/.style = {
     draw=black, thick,
    circle, inner sep=2pt
  },
  arrow/.style = {thick, -Stealth, color=black, draw=gray!30!black},
  group/.style = {
    draw=gray!10!black, dotted,
    inner sep=8pt, rectangle, rounded corners
  },
}

\begin{document}

\newgeometry{top=1in,left=0.75in,right=0.75in,bottom=0.75in}

\title{SmoothRot: Combining Channel-Wise Scaling and Rotation for Quantization-Friendly LLMs}

\author{%
  \begin{minipage}[t]{0.32\linewidth}
    \centering
    Patrik Czakó\\[0.5ex]
    \textit{Doctoral School of Applied Informatics and Applied Mathematics, Obuda University}\\
    Budapest, Hungary\\
    czako.patrik@stud.uni-obuda.hu
  \end{minipage}\hfill
  \begin{minipage}[t]{0.32\linewidth}
    \centering
    Gábor Kertész\\[0.5ex]
    \textit{John von Neumann Faculty of Informatics, Obuda University}\\
    Budapest, Hungary\\
    kertesz.gabor@nik.uni-obuda.hu
  \end{minipage}\hfill
  \begin{minipage}[t]{0.32\linewidth}
    \centering
    Sándor Szénási\\[0.5ex]
    \textit{John von Neumann Faculty of Informatics, Obuda University}\\
    Budapest, Hungary\\
    szenasi.sandor@nik.uni-obuda.hu
  \end{minipage}%
}

\maketitle

\begin{abstract}
We present SmoothRot, a novel post-training quantization technique to enhance the efficiency of 4-bit quantization in Large Language Models (LLMs). SmoothRot addresses the critical challenge of massive activation outliers, by integrating channel-wise scaling with Hadamard transformations. Our technique effectively transforms extreme outliers into quantization-friendly activations, significantly improving quantization accuracy. Experiments conducted on popular LLMs (LLaMA2 7B, LLaMA3.1 8B, and Mistral 7B) demonstrate that SmoothRot consistently reduces the performance gap between quantized and FP16 models by approximately 10-30\% across language generation and zero-shot reasoning tasks, without introducing additional inference latency. Code is available at \url{https://github.com/czakop/smoothrot}
\end{abstract}


\section{Introduction}

Large Language Models (LLMs)~\cite{touvronLlama2Open2023, grattafioriLlama3Herd2024, jiangMistral7B2023}  have shown remarkable capabilities in natural language processing, becoming central to many artificial intelligence applications. However the rapid increase in models sizes required to achieve these impressive results has significantly raised their training and inference costs in terms of time, memory and energy consumption compared to smaller models~\cite{yangTrainingExperimentalLanguage2023}. Consequently, efficient post-training compression techniques are essential for practical LLM deployment. Among these, Post-Training Quantization (PTQ) stands out, achieving state-of-the-art results by minimizing performance loss at given compression ratios \cite{wangModelCompressionEfficient2024}.

Efficient inference optimization requires quantizing both weights (trained parameters) and activations (intermediate computational results) \cite{jacobQuantizationTrainingNeural2017}. While weight-only quantization can achieve extremely low precision (even down to 1-2 bits) with minimal performance loss \cite{kimSqueezeLLMDenseandSparseQuantization2024, tsengQuIPEvenBetter2024}, activation quantization is more challenging due to the presence of significant outlier values. Although recent research efforts address these activation outliers \cite{czako2025activation}, effectively reducing their impact remains a major research challenge.

Building upon the recent finding in \cite{czako2025turning} that channel-wise scaling~\cite{xiaoSmoothQuantAccurateEfficient2023} applied before rotation~\cite{ashkboosQuaRotOutlierFree4Bit2024} significantly reduces layer-wise quantization errors caused by massive activation outliers, we propose SmoothRot, a novel PTQ technique that combines channel-wise scaling and rotation to produce activations that are more amenable to quantization. Our main contributions are:

\begin{itemize}
    \item We demonstrate the efficient integration of channel-wise scaling into a rotation-based quantization framework~\cite{ashkboosQuaRotOutlierFree4Bit2024}, introducing no additional inference latency.
    \item We extend prior layer-wise observations from \cite{czako2025turning} to end-to-end model performance, showing SmoothRot consistently outperforms the rotation-only method in the 4-bit quantization scenario by reducing the performance gap to FP16 models by approximately $10-30\%$ across multiple tasks.
    \item We provide an in-depth analysis of hyperparameter selection, including the impact of the calibration dataset and migration strength on quantization performance.
    \item We show SmoothRot's compatiblity with advanced weight quantization and rotation techniques by presenting experimental results using GPTQ~\cite{frantarOPTQACCURATEPOSTTRAINING2023} and SpinQuant~\cite{liuSpinQuantLLMQuantization2024}.
\end{itemize}
\section{Background}

\begin{figure*}[t!]
    \centering
    \subfigure[Original]{
        \includegraphics[width=0.3\linewidth]{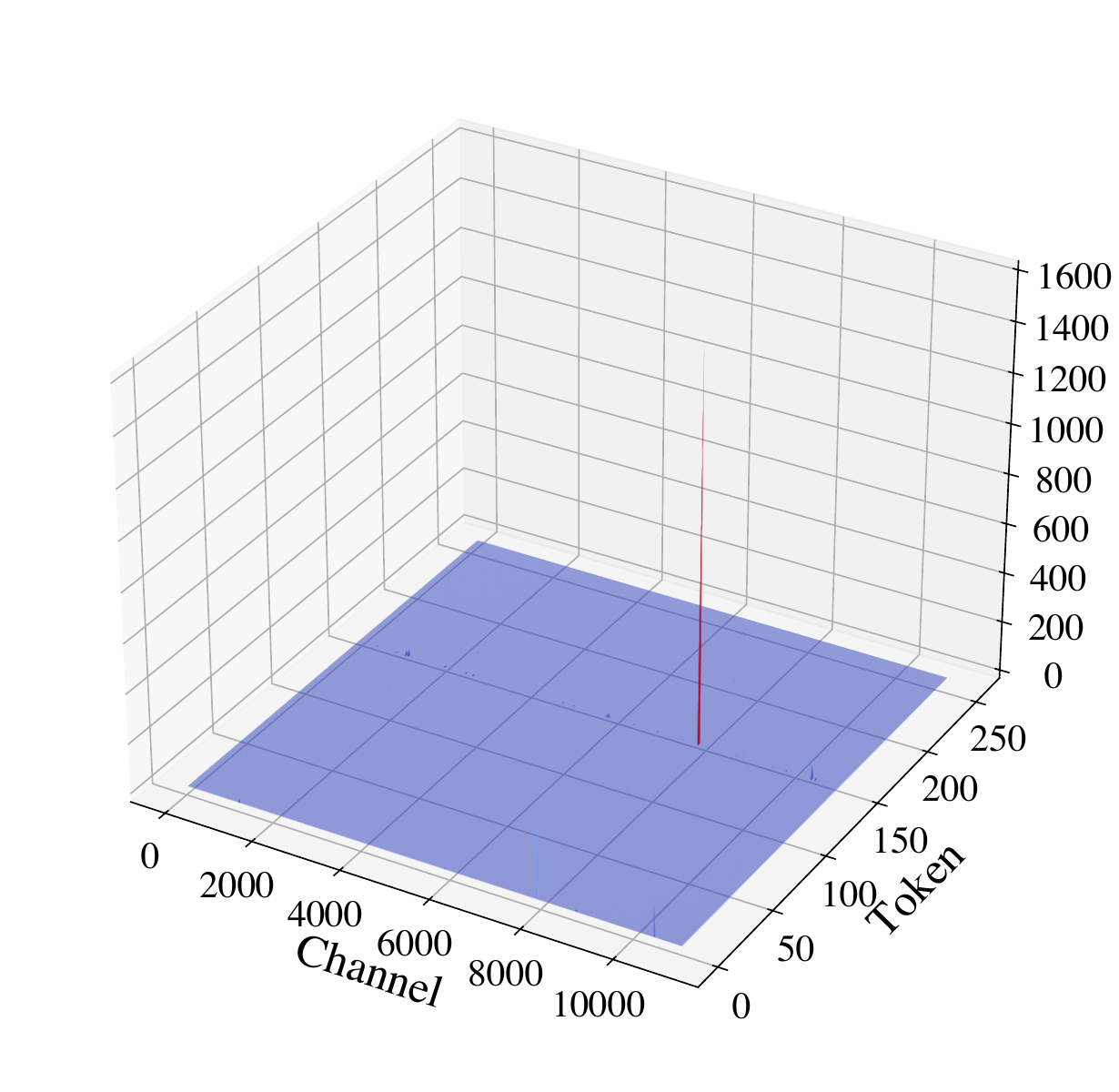}
        \label{fig:act_orig}
    }
    \subfigure[QuaRot]{
        \includegraphics[width=0.3\linewidth]{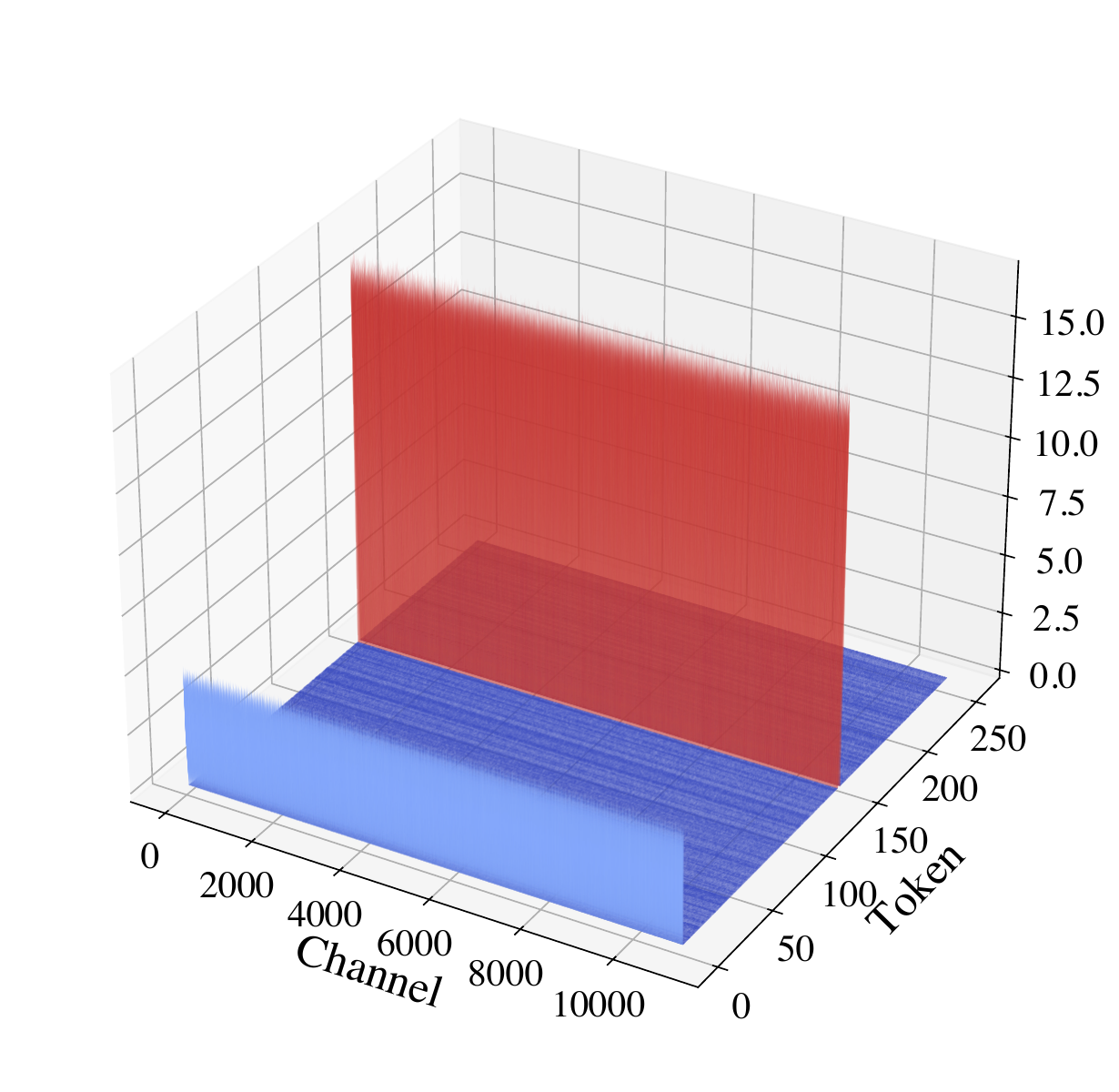}
        \label{fig:act_rot}
    }
    \subfigure[SmoothRot]{
        \includegraphics[width=0.3\linewidth]{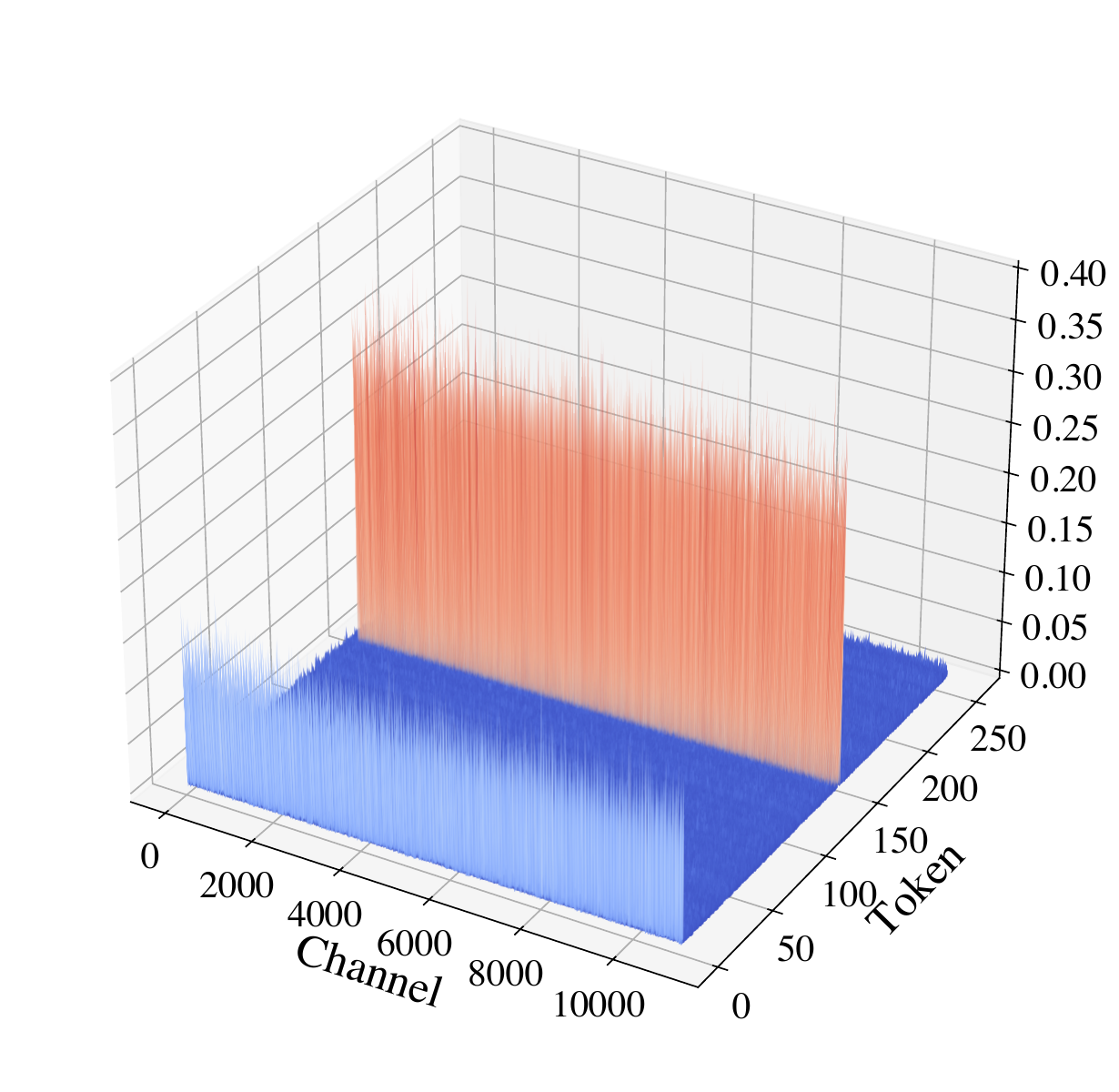}
        \label{fig:act_srot}
    }
    \vspace{-5pt}
    \caption{Input activation magnitudes of the second down-projection layer of LLaMA2 7B under different quantization schemes. The original model \subref{fig:act_orig} shows untransformed activations exhibinting massive outliers. QuaRot \subref{fig:act_rot} rotates the activations using a Hadamard transformation, leading to significantly smaller magntidues. SmoothRot \subref{fig:act_srot} further suppresses outliers by applying an additional channel-wise scaling operation before rotation, resulting in the most quantization-friendly activation distribution.}
    \label{fig:act}
    \vspace{-0.1in}
\end{figure*}

Quantization reduces the precision of weights and activations in neural networks to decrease memory usage and latency \cite{jacobQuantizationTrainingNeural2017}. The integer quantization process can be mathematically represented as:

\begin{equation}
    \mathbf{X}_{INT} = \text{clamp}(\lfloor \frac{\mathbf{X}}{\Delta} \rceil + z, 0, 2^b -1).
\end{equation}

The quantization step size ($\Delta$) and zero-point ($z$) are computed using $\Delta = \max(\mathbf{X})/(2^{b-1} -1)$, $z = 0$ for symmetric quantization, and $\Delta = (\max(\mathbf{X}) - \min(\mathbf{X}))/(2^b -1)$, $z = -\lfloor \min(\mathbf{X})/\Delta\rceil$ for asymmetric quantization. In these formulas, $b$ represents the bit-width, $\mathbf{X}_{INT}$ denotes the quantized (integer) tensor, $\mathbf{X}$ represents the original (floating-point) tensor, and $\lfloor\cdot\rceil$ indicates the rounding function. The input tensor is then dequantized as $Q(\mathbf{X}) = (\mathbf{X}_{INT} - z) \cdot \Delta$. The reconstruction error introduced by this process primarily depends on the range of the tensor values. Notably, in LLMs, this range can significantly increase due to outliers, which are typically classified into systematic outliers~\cite{dettmersLLMint88bitMatrix2022, xiaoSmoothQuantAccurateEfficient2023} and massive outliers~\cite{sunMassiveActivationsLarge2024}. In this paper, we specifically address the latter category, described next.

\subsection{Massive Activation Outliers}

Massive activations in LLMs \cite{sunMassiveActivationsLarge2024} are defined as activations whose magnitude surpasses 100 and are approximately 1000 times larger than the median magnitude of their hidden states. Although these activations are present in the hidden states of nearly all decoder layers, they are largely suppressed by the input layer normalizations before quantization occurs at the Key, Value, and Query projection layers, thus mitigating much of their adverse impact on quantization error. However, recent studies \cite{yangMitigatingQuantizationErrors2024, sonPrefixingAttentionSinks2024} have shown that in Gated Linear Unit (GLU) architectures, spikes occur in the input activations of down-projection layers withing Feed-Forward Network (FFN) modules, leading to significant quantizations errors \cite{czako2025turning}. As depicted in \cref{fig:act_orig}, these outliers typically occur in specific channels and are associated with a small subset of tokens, often those with low semantic value (e.g., periods, new lines, etc.).

\subsection{Outlier Mitigation Techniques}

Several approaches have been proposed to mitigate the challenges that activation outliers pose for quantization \cite{czako2025activation}. While mixed-precision techniques \cite{yangMitigatingQuantizationErrors2024,sonPrefixingAttentionSinks2024} have demonstrated promise using higher bit-width for tokens with massive outliers, they do not fully exploit the advantages of low-bit integer quantization targeted in this study, and thus are excluded from our scope.

A more effective alternative is the application of equivalent linear transformations, where an invertible matrix $\mathbf{A} \in \mathbb{R}^{c_{in} \times c_{in}}$ transforms the input tensor $\mathbf{X}$ into $\hat{\mathbf{X}} = \mathbf{XA}$, making quantization simpler. The weight tensor $\mathbf{W}$ is inversely transformed to maintain the original layer output $\mathbf{Y}$:

\begin{equation} \label{eq:eq_transformation}
\mathbf{Y} = \mathbf{X} \mathbf{W} = \mathbf{X}\underbrace{\left( \mathbf{A}\mathbf{A}^{-1}\right)}_{\mathbb{I}} \mathbf{W} = \underbrace{\left(\mathbf{X} \mathbf{A}\right)}_{\hat{\mathbf{X}}} \cdot \underbrace{\left(\mathbf{A}^{-1} \mathbf{W}\right)}_{\hat{\mathbf{W}}}.
\end{equation}

Here, $\mathbb{I} \in \mathbb{R}^{c_{in} \times c_{in}}$ denotes the identity matrix of size equals to the number of input channels. Although some methods \cite{sunFlatQuantFlatnessMatters2024,maAFFINEQUANTAFFINETRANSFORMATION2024} directly optimize the transformation matrix $\mathbf{A}$, two simplified yet widely adopted approaches impose specific constraints on $\mathbf{A}$. SmoothQuant~\cite{xiaoSmoothQuantAccurateEfficient2023} employs channel-wise scaling, constructing $\mathbf{A}$ as a diagonal matrix from a scaling factor $\mathbf{s} \in \mathbb{R}^{c_{in}}$, such that $\mathbf{A}^{-1} = \mathrm{diag}(\mathbf{s})$. This technique shifts quantization difficulty from activations to layer weights, effectively addressing systematic outliers, but it struggles alone to adequately handle massive outliers, particularly in low-bit quantizations scenarios.

Conversely, QuaRot~\cite{ashkboosQuaRotOutlierFree4Bit2024} proposes rotation activations using an orthogonal matrix in \cref{eq:eq_transformation}. As illustrated in \cref{fig:act_rot}, rotation evenly distributes outlier values across channels, proving superior to channel-wise scaling for massive outliers. Additionally, rotation effectively reduces quantization errors in low-bit settings, leading researchers to explore this approach further in later studies ~\cite{liuSpinQuantLLMQuantization2024, linDuQuantDistributingOutliers2024, xiangDFRotAchievingOutlierFree2024}.

However, \cite{czako2025turning} indicates that rotation alone can amplify quantization errors in 4-bit scenarios involving extremely large outliers. The study recommends initially applying smoothing to partially transfer quantization difficulty to weights, followed by rotation, which redistributes the complexity across channels. In this paper, we adopt and enhance this combined approach, augmenting QuaRot's end-to-end quantization framework, which we term SmoothRot.

\section{Methodology}

\begin{figure*}[t!]
    \centering
    \scalebox{0.9}{\begin{tikzpicture}[node distance=0.25cm and 0.5cm]
  \node           (input)  {$\mathbf{XQ}$};
  \node[nonlinearity, right=of input, draw=colRot]  (norm1)  {$\frac{x}{\|x\|}$};
  \node[smallblock, right=of norm1, dashed, draw=colRot] (quant1) {\rotatebox{90}{quantize}};
  \coordinate      (split) at ($(quant1.east)+(2.5em,0)$);
  \node[block,     above right=of split, yshift=0.5em, label=above:gate-projection] (Wgate) {$\mathbf{Q}^\top\mathbf{\Gamma W}_\text{gate}$};
  \node[block, below right=of split, yshift=-0.5em, label=above:up-projection] (Wup)   {$\mathbf{Q}^\top\mathbf{\Gamma}\mathbf{W}_\text{up}\mathbf{\Lambda}^{-1}$};
  \node[nonlinearity, right=2.5em of Wgate] (sigma) {$\sigma$};
  \node[operator,  below right=of sigma, yshift=-0em, draw=colSmooth] (mult) {$\times$};
  \node[smallblock,     right=of mult, draw=colSmoothRot]       (hadamard) {\rotatebox{90}{hadamard}};
  \node[smallblock,     right=1.5em of hadamard, dashed, draw=colSmoothRot] (quant2) {\rotatebox{90}{quantize}};
  \node[block,     right=2em of quant2, label=above:down-projection] (Wdown) {$\mathbf{H\Lambda W}_\text{down}\,\mathbf{Q}$};
  \node   (out)    [right=of Wdown] {$\mathbf{YQ}$};

  \draw[arrow, draw=colRot]  (input)  -- (norm1);
  \draw[arrow, draw=colRot]  (norm1)  -- (quant1);
  \draw[arrow, draw=colRot, -, dashed]  (quant1) -- (split);
  \draw[arrow, draw=colRot, dashed]      (split)  |- (Wgate);
  \draw[arrow, draw=colRot, dashed]      (split)  |- (Wup);
  \draw[arrow] (Wgate)  -- (sigma);
  \draw[arrow]   (sigma)  -| (mult);
  \draw[arrow, draw=colSmooth] (Wup)    -| (mult);
  \draw[arrow, draw=colSmooth](mult)   -- (hadamard);
  \draw[arrow, draw=colSmoothRot]  (hadamard)-- (quant2);
  \draw[arrow, draw=colSmoothRot, dashed]  (quant2) -- (Wdown);
  \draw[arrow, draw=colRot]  (Wdown)  -- (out);

  \node[group, label=above:RMSNorm]   (g1)  [fit=(norm1)(quant1)] {};
  \node[above=0.5em of Wgate]         (aux) {};
  \node[group, label=above:FFN]       (g2)  [fit=(split)(Wgate)(sigma)(Wup)(mult)(Wdown)(aux)] {};

  \matrix[
    draw, fill=white,
    anchor=south east,
    font=\scriptsize,
    matrix of nodes,
    nodes={inner sep=1pt, anchor=west},
    column sep=4pt, row sep=2pt,
  ] at ([xshift=-0.0em] g2.south east) {
    \tikz{\draw[arrow, dashed]   (0,0) -- (0.4,0);} & Quantized \\
    \tikz{\draw[arrow]   (0,0) -- (0.4,0);} & Untransformed \\
    \tikz{\draw[arrow, colRot]       (0,0) -- (0.4,0);} & Rotated       \\
    \tikz{\draw[arrow, colSmooth]    (0,0) -- (0.4,0);} & Smoothened    \\
    \tikz{\draw[arrow, colSmoothRot] (0,0) -- (0.4,0);} & SmoothRot     \\
  };
\end{tikzpicture}}
    \caption{SmoothRot applied to a LLaMA-style FFN. Following QuaRot \cite{ashkboosQuaRotOutlierFree4Bit2024}, a diagonal matrix ($\mathbf{\Gamma}$) consisting of the RMSNorm scaling parameters is absorbed into the weight matrices $\mathbf{W}_\text{gate}$ and $\mathbf{W}_\text{up}$, ensuring computational invariance during the rotation with $\mathbf{Q}$. An online Hadamard transformation is applied to the input activations of the down-projection, which have been smoothed by the matrix $\mathbf{\Lambda}$ absorbed into the weights of the up-projection layer, resulting in a more quantization-friendly activation tensor for subsequent quantization.}
    \label{fig:arch}
\end{figure*}
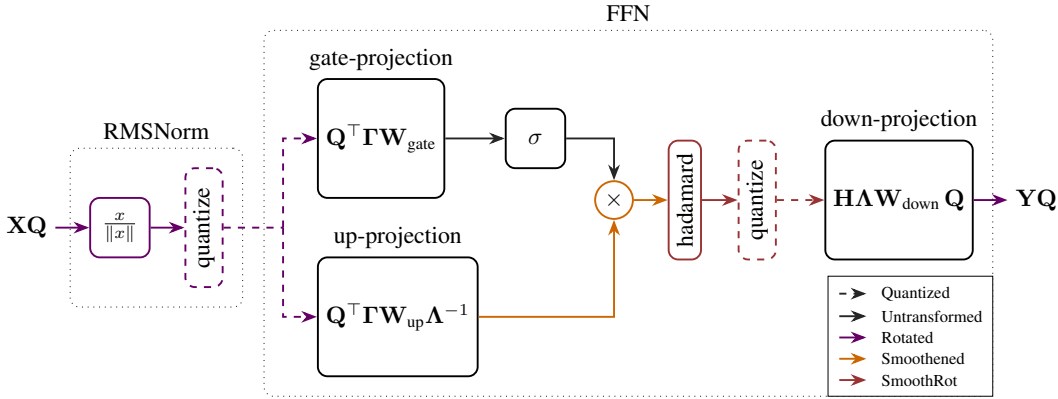

SmoothRot extends the quantization framework of QuaRot~\cite{ashkboosQuaRotOutlierFree4Bit2024} by introducing an additional smoothing stage, as proposed in \cite{czako2025turning}. This modification exclusively targets the FFN modules, as illustrated in \cref{fig:arch}. Initially, channel-wise scaling is applied to the input activations of the down-projection layers. Following the methodology from \cite{xiaoSmoothQuantAccurateEfficient2023}, scaling factors are calibrated offline using a calibration dataset with the formula $s_j = \max\left(\lvert \mathbf{X}_j \rvert\right)^\alpha / \max\left(\lvert \mathbf{W}_j \rvert\right)^{1-\alpha}$, where the hyperparameter $\alpha$ controls the migrations strength. These factors are absorbed into the weights of the up-projection layers as $\mathbf{\Lambda}=\mathrm{diag}(\mathbf{s})$, leveraging the channel-independence property of the subsequent Hadamard product, ensuring no additional latency during inference. An inverse scaling transformation is simultaneously applied to the down-projection layer weights.

Subsequently, the model weights are rotated using random Hadamard matrices $\mathbf{Q}$, accompanied by two additional Hadamard transformations integrated into the forward pass, as described in \cite{ashkboosQuaRotOutlierFree4Bit2024}. One transformation is applied immediately before the down-projection layer, transforming the smoothened activations, resulting in the final transformation matrix $\mathbf{A}=\left(\mathbf{H\Lambda}\right)^{-1}$ in \eqref{eq:eq_transformation}. To maintain computational invariance, linear components ($\mathbf{\Gamma}$) from RMSNorm are fused with adjacent weight matrices. This approach preserves the commutative property $\RMSNorm(\mathbf{X}) = \RMSNorm(\mathbf{XQ}^\top)\mathbf{Q}$. Although \cite{czako2025turning} recommends smoothing rotations only for down-projection layers due to limited gains elsewhere, it is practically challenging to apply smoothing directly before module inputs because of the modifications to RMSNorm. Specifically, the approach in \cite{xiaoSmoothQuantAccurateEfficient2023}, which fuses scaling factors into preceding normalization layers, is incompatible with our modification.

Finally, weight quantization is performed using either GPTQ~\cite{frantarOPTQACCURATEPOSTTRAINING2023} or the Round-To-Nearest (RTN) method, while online quantization operations are integrated into the forward pass for both activations and the Key-Value (KV) cache.
\section{Evaluation}

\begin{table}[!t]
\centering
\caption{Wikitext-2 and C4 Perplexity Results ($\downarrow$) of 4-bit Weight-Activation Quantized Models}
\label{table:main_ppl}
\resizebox{0.85\columnwidth}{!}{
\begin{tabular}{lcccccc}
\toprule
\multirow{2}{*}{\textbf{Method}} & \multicolumn{2}{c}{LLaMA2 7B} & \multicolumn{2}{c}{LLaMA3.1 8B} & \multicolumn{2}{c}{MistralV3 7B} \\ \cmidrule(l{2pt}r{2pt}){2-3} \cmidrule(l{2pt}r{2pt}){4-5} \cmidrule(l{2pt}r{2pt}){6-7}
            & Wiki & C4 & Wiki & C4 & Wiki & C4 \\ \midrule
FP16        & 5.47 & 7.26 & 6.24 & 9.54 & 5.32 & 8.47 \\ \midrule
QuaRot      & 8.33 & 11.69& 9.83 & 15.78& 6.41 & 10.01 \\
SmoothRot   & \textbf{7.51} & \textbf{10.47}& \textbf{9.48} & \textbf{15.02}& \textbf{6.19} & \textbf{9.78} \\ 
\bottomrule
\end{tabular}
}
\end{table}
\begin{table*}[!t]
\centering
\caption{Zero-Shot QA Task Results ($\%$) of 4-bit Weight-Activation Quantized Models}
\label{table:main_acc}
\resizebox{0.7\linewidth}{!}{
\begin{tabular}{llccccccc}
\toprule
\textbf{Model} & \textbf{Method} & \textbf{ARC-C} &  \textbf{ARC-E} & \textbf{HellaSwag} & \textbf{LAMBADA} & \textbf{PICA} & \textbf{WinoGrande} & \textbf{Avg.} \\
\midrule
\multirow{3}{*}{LLaMA2 7B}     & FP16        & 46.33 & 74.62 & 75.99 & 73.92 & 79.00 & 69.14 & 69.61 \\ \cmidrule(lr){2-9} 
                                & QuaRot      & 35.41 & 59.43 & 65.14 & 58.22 & 73.23 & 59.91 & 55.94 \\
                                & SmoothRot   & 38.57 & 65.32 & 67.82 & 62.72 & 74.97 & 61.80 & \textbf{59.83} \\ \midrule
\multirow{3}{*}{LLaMA3.1 8B}    & FP16        & 53.58 & 81.23 & 78.90 & 75.82 & 81.28 & 73.64 & 73.12 \\ \cmidrule(lr){2-9} 
                                & QuaRot      & 40.96 & 67.34 & 68.75 & 58.24 & 74.81 & 66.38 & 60.69 \\
                                & SmoothRot   & 41.98 & 67.85 & 70.59 & 64.51 & 74.86 & 65.27 & \textbf{62.88} \\ \midrule
\multirow{3}{*}{MistralV3 7B}   & FP16        & 52.30 & 78.20 & 80.42 & 75.33 & 82.26 & 73.88 & 73.08 \\ \cmidrule(lr){2-9} 
                                & QuaRot      & 45.22 & 72.26 & 75.14 & 69.40 & 79.22 & 68.82 & 66.86 \\
                                & SmoothRot   & 46.59 & 73.19 & 75.97 & 70.33 & 79.11 & 68.11 & \textbf{67.47} \\ \bottomrule
\end{tabular}
}
\end{table*}
\begin{table}[!t]
\centering
\caption{Perplexity Results ($\downarrow$) of 4-bit Weight-Activation Quantized LLaMA2 7B Using Different Calibration Datasets}
\label{table:calib}
\resizebox{0.6\columnwidth}{!}{
\begin{tabular}{lccc}
\toprule
\multirow{2}{*}{\textbf{Calibration set}} & \multicolumn{3}{c}{\makecell{\textbf{PPL ($\downarrow$)}}} \\ \cmidrule(lr){2-4}
        & Wiki & C4 & PTB \\ \midrule
Wiki    & 7.51 & 10.47 & 63.96 \\
C4      & 7.49 & 10.45 & 64.32 \\
PTB     & 7.53 & 10.50 & 62.68 \\
Random  & 7.75 & 10.78 & 64.06 \\ \midrule
- (QuaRot) & 8.33 & 11.69 & 76.61 \\
\bottomrule
\end{tabular}
}
\end{table}
\begin{figure}[t!]
    \centering
    \includegraphics[width=0.8\columnwidth]{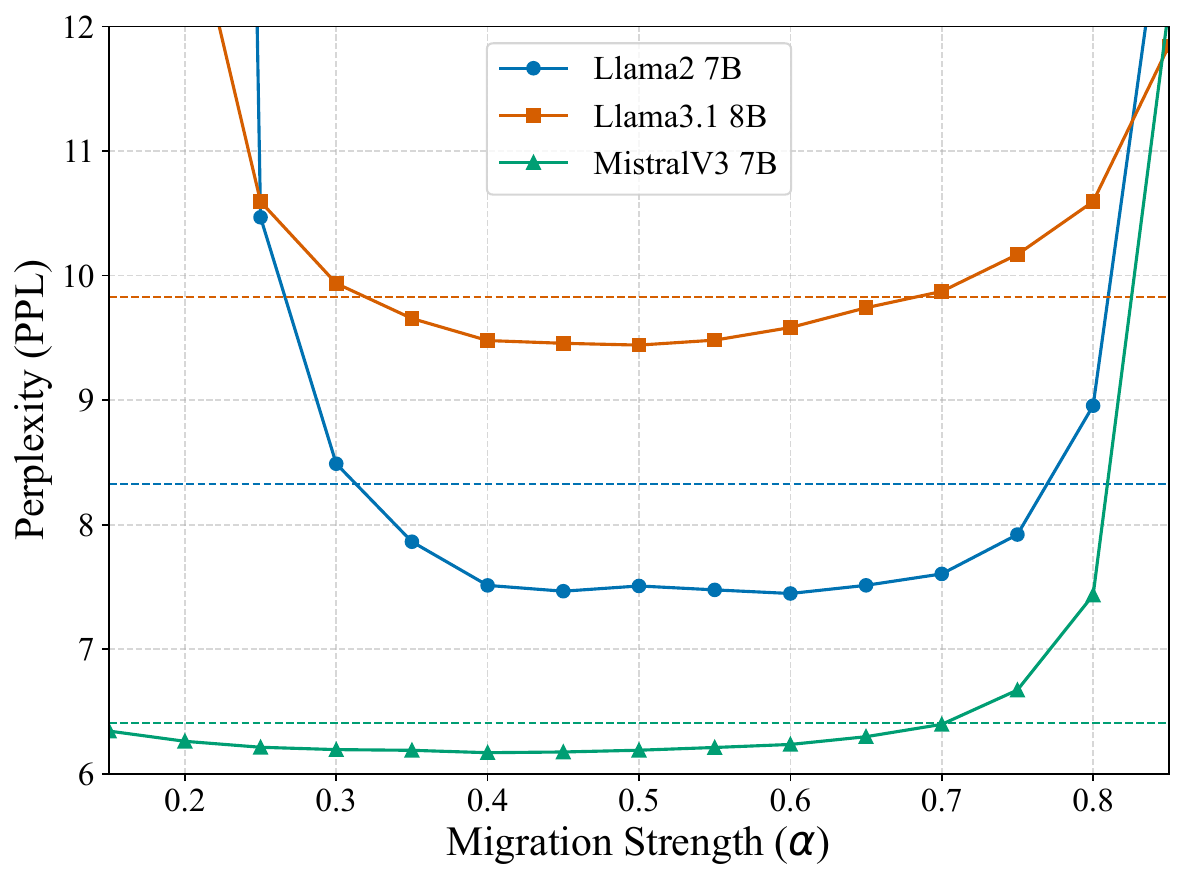}
    \caption{Effect of migration strength ($\alpha$) on Wikitext-2 perplexity. The dashed horizontal lines indicate the perplexity achieved by QuaRot~\cite{ashkboosQuaRotOutlierFree4Bit2024} for each model. Lower perplexity values correspond to better language modeling quality.}
    \label{fig:alpha}
    \vspace{-0.2in}
\end{figure}
\begin{table}[!t]
\centering
\caption{Evaluation Results of 4-bit Weight-Activation Quantized Models Using GPTQ Weight Quantization}
\label{table:gptq}
\resizebox{0.95\columnwidth}{!}{
\begin{tabular}{llccc}
\toprule
\textbf{Model} & \textbf{Method} & \textbf{Wiki ($\downarrow$)} & \textbf{C4 ($\downarrow$)} & \textbf{0-shot Avg. ($\uparrow$)} \\ \midrule
\multirow{3}{*}{LLaMA2 7B}     & FP16        & 5.47 & 7.26 & 69.61 \\ \cmidrule(lr){2-5} 
                                & QuaRot      & 6.11 & 8.31 & 65.32 \\
                                & SmoothRot   & 6.11 & 8.42 & 65.64 \\ \midrule
\multirow{3}{*}{LLaMA3.1 8B}   & FP16        & 6.24 & 9.54 & 73.12 \\ \cmidrule(lr){2-5} 
                                & QuaRot      & 8.12 & 13.16& 65.65 \\
                                & SmoothRot   & 8.24 & 13.33& 65.43 \\ \midrule
\multirow{3}{*}{MistralV3 7B}   & FP16        & 5.32 & 8.47 & 73.08 \\ \cmidrule(lr){2-5} 
                                & QuaRot      & 5.79 & 9.19 & 69.51 \\
                                & SmoothRot   & 5.81 & 9.24 & 69.45 \\ 
\bottomrule
\end{tabular}
}
\end{table}

\subsection{Experimental settings}

\textbf{Tasks and Baselines.} We evaluate SmoothRot on three models: LLaMA2 7B~\cite{touvronLlama2Open2023}, LLaMA3.1 8B~\cite{grattafioriLlama3Herd2024} and Mistral 7B~\cite{jiangMistral7B2023}. Our evaluations cover language generation tasks using perplexity scores (PPL) on Wikitext-2~\cite{merityPointerSentinelMixture2017} and C4~\cite{raffelExploringLimitsTransfer2023} datasets, an accuracy on six zero-shot commonsense reasoning tasks including PIQA~\cite{biskPIQAReasoningPhysical2019}, WinoGrande~\cite{sakaguchiWinoGrandeAdversarialWinograd2019}, HellaSwag~\cite{zellersHellaSwagCanMachine2019}, LAMBADA~\cite{papernoLAMBADADatasetWord2016}, Arc-Easy and Arc-Challenge~\cite{clarkThinkYouHave2018}. Evaluations are conducted using the LM Evaluation Harness~\cite{eval-harness} with default parameters. Our main results compare SmoothRot against QuaRot~\cite{ashkboosQuaRotOutlierFree4Bit2024}, highlighting the improvements achieved by our technique. Additionally, SpinQuant~\cite{liuSpinQuantLLMQuantization2024} is included for comparison in ablation studies. For consistency, all reported results are reproduced by us.

\textbf{Implementation Details.} We implemented SmoothRot based on the frameworks provided by SmoothQuant and QuaRot. The smoothing factors are calibrated offline using 512 sentences (512 tokens each) randomly selected from Wikitext-2, completed within just a few seconds on a single NVIDIA A100 GPU. The optimal migration strength ($\alpha$) is determined individually performing a linear search per model: 0.6 (LLaMA2 7B), 0.5 (LLaMA3.1 8B), and 0.45 (Mistral 7B). The effects of calibration dataset choice and migration strength are explored further in our ablation studies. Rotation is performed with random Hadamard matrices offline and exact matrices online. SmoothRot adds no latency overhead compared to QuaRot, benefiting from efficient kernel implementations due to fused transformation matrices.

\textbf{Quantization}. All model components are quantized to 4 bits, following the settings described in \cite{ashkboosQuaRotOutlierFree4Bit2024}. Specifically, activations use per-token symmetric quantization with a constant clipping ratio of 0.9. Keys and Values use asymmetric quantization with a group size of 128 and a constant clipping ratio of 0.95. Weights are primarily quantized using the RTN method with per-channel symmetric quantization, with clipping ratios determined through a linear search over the squared error. GPTQ~\cite{frantarOPTQACCURATEPOSTTRAINING2023} weight quantization is also explored in ablation studies, where use 128 samples from WikiText-2 training set with 2048 sequence length as the calibration set.

\subsection{Main Results}

\textbf{Language Generation Tasks.} As shown in \cref{table:main_ppl} SmoothRot consistently outperforms QuaRot on both Wikitext-2 and C4 datasets across all models. Notably, SmoothRot reduces the perplexity by $0.82$ for LLaMA2 7B on Wikitext-2, narrowing the gap to the FP16 baseline by nearly $30\%$.

\textbf{Zero-Shot Tasks.} \cref{table:main_acc} displays SmoothRot's accuracy on the six evaluated zero-shot commonsense tasks. Our approach outperforms QuaRot across nearly all tasks and models, partially recovering lost accuracy compared to the FP16 baseline. The most significant improvements appear with LLaMA2 7B, achieving an average improvement of $3.89\%$, while improvements for LLaMA3.1 8B and Mistral 7B are $2.19\%$ and $0.61\%$, respectively.

\subsection{Ablation Studies}

\textbf{Choice of Calibration Set.} \cref{table:calib} shows results using different calibration datasets: Wikitext-2, C4, Penn TreeBank (PTB) \cite{marcusBuildingLargeAnnotated1993}, and random tokens. We observe minimal performance variation across calibration sets, and notably, random token calibration still significantly outperforms QuaRot.

\textbf{Effect of Migration Strength.} \cref{fig:alpha} demonstrates how migration strength ($\alpha$) affects model perplexity. Optimal results generally occur around $\alpha=0.5$. Although LLaMA models only achieve better results than QuaRot within an $\alpha$ range of $0.35-0.7$ to outperform QuaRot, Mistral achieves improvements even at $\alpha$ as low as $0.15$.

\textbf{GPTQ Weight Quantization.} \cref{table:gptq} compares SmoothRot with QuaRot when GPTQ~\cite{frantarOPTQACCURATEPOSTTRAINING2023} is used for weight quantization instead of the simple RTN method. Interestingly, in this case, the improvements offered by SmoothRot over QuaRot diminish, with results becoming very close or occasionally slightly worse. This outcome suggests that further investigation is needed to understand the interaction between SmoothRot and advanced weight quantization techniques like GPTQ.

\begin{table}[!t]
\centering
\caption{Evaluation Results of 4-bit Weight-Activation Quantized LLaMA2 7B Using SpinQuant Rotation Matrices}
\label{table:spinquant}
\resizebox{0.7\columnwidth}{!}{
\begin{tabular}{lccc}
\toprule
\textbf{Method} & \textbf{Wiki ($\downarrow$)} & \textbf{C4 ($\downarrow$)} & \textbf{0-shot Avg. ($\uparrow$)} \\ \midrule
FP16        & 5.47 & 7.26 & 69.61 \\ \midrule
SpinQuant   & 6.32 & 9.47 & 61.68 \\
SmoothRot   & 6.47 & 9.56 & 61.13 \\
\bottomrule
\end{tabular}
}
\end{table}
\textbf{Compatibility with SpinQuant.} Similar to QuaRot, SpinQuant~\cite{liuSpinQuantLLMQuantization2024} rotates input activations to reduce quantization error. Although it replaces the random Hadamard matrices with optimized orthogonal matrices, it still applies exact Hadamard transformation on the input of down-projection layers. This leaves room for SmoothRot to adjust these activations by applying channel-wise scaling beforehand. We evaluate SmoothRot combined with SpinQuant's rotation matrices on LLaMA2 7B, using optimized matrices published by \cite{liuSpinQuantLLMQuantization2024} and a migration strength of $0.5$. As reported in \cref{table:spinquant}, incorporating channel-wise scaling before the online Hadamard transformation results in slightly worse outcomes compared to SpinQuant alone. However, we believe this setup warrants further investigation, and optimizing the rotation matrices after applying smoothing may lead to improved performance.

\section{Conclusion}

In this paper, we revisit pre-quantization transformations to address activation outliers in LLMs, aiming to improve 4-bit quantization of both weights and activations. Specifically, we target massive outliers predominantly observed in the inputs of down-projection layers. We introduce SmoothRot, a highly efficient post-trainig quantization technique that significantly extends QuaRot by combining its rotation-only quantization framework with channel-wise scaling. SmoothRot effectively transforms extreme outliers, often exceeding magnitudes of $1000$, into quantization-friendly activations below $0.5$, achieving over a $30$-fold improvement compared to QuaRot alone. As a result, SmoothRot consistently reduces the performance gap to unquantized models by approximately $10-30\%$ across language generation and zero-shot tasks under the W4A4KV4 RTN quantization setting, without introducing any additional latency overhead.

\section*{Acknowledgment}
We thank the Doctoral School of Applied Informatics and Applied Mathematics at Obuda University for their valuable support. We also acknowledge Digital Government Development and Project Management Ltd. for granting us access to the Komondor HPC facility in Hungary. Finally, we extend our sincere gratitude to Attila Farkas for his assistance with the experimental setup.

\printbibliography

\end{document}